\documentclass[10pt,twocolumn,letterpaper]{article}

\usepackage[pagenumbers]{cvpr} 

\usepackage[T1]{fontenc}
\usepackage{amsmath,amssymb,booktabs,graphicx,multirow,array}

\definecolor{cvprblue}{rgb}{0.21,0.49,0.74}
\usepackage[pagebackref,breaklinks,colorlinks,allcolors=cvprblue]{hyperref}

\def\confName{CVPR}
\def\confYear{2027}

\title{Towards High-Fidelity Learned Video Compression}

\author{Bumyoon Kim \quad Yongseong Kim \quad Inhyuk Jeong \quad Byeungwoo Jeon\\
Department of Electrical and Computer Engineering, Sungkyunkwan University
}

\begin{document}
\maketitle
\begin{abstract}
Professional video workflows operate at qualities that learned video codecs do not reach. We measure how much of a professional quality range a codec covers, from the lowest quality of the standard test conditions, VVC at QP 37, to the highest 4:2:2 grade of the production codec ProRes. To compare inter-frame coding alone, every codec receives original frames as references at the intra positions. On 10-bit YCbCr 4:2:2 test sequences, four recent learned codecs cover 24 to 38\% of this range. Their encoders are feed-forward networks with a fixed latent capacity, the number of latent values per sample, of at most 0.25, and removing their quantization raises the quality by at most 1\,dB. We propose a learned inter-frame codec whose encoder searches, as the encoders of conventional codecs do. It searches block motion, which a learned network refines, and it searches among learned block transforms whose latent capacity ranges up to 1, selecting the partition, the skip decisions, and the quantized coefficients by a rate-distortion cost. Our codec reaches the upper end of the range on all six sequences. With all residual blocks fixed to 16$\times$16, whose latent capacity is 0.25, the same codec covers 55\% of the range, with or without quantization.
\end{abstract}

\section{Introduction}
\label{sec:intro}

\begin{figure}[t]
 \centering
 \includegraphics[width=\linewidth]{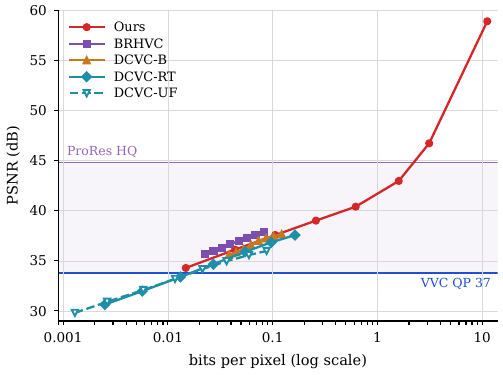}
 \caption{Rate-distortion curves averaged over six JVET 4:2:2 sequences. The band marks the comparison range, from VVC at QP 37 to ProRes HQ. Four learned codecs stop near 38\,dB, whereas the proposed codec reaches the upper end of the range.}
 \label{fig:teaser}
\end{figure}

Video compression removes spatial and temporal redundancy so that a frame can be reconstructed at a target quality from as few bits as possible. Conventional codecs such as HEVC~\cite{tcsvt2012hevc} and VVC~\cite{tcsvt2021vvcoverview} are block-based hybrid systems whose coding tools are selected, block by block, by an encoder-side candidate search that minimizes a rate-distortion cost. Learned video compression instead obtains transforms, probability models, and temporal processing from data~\cite{cvpr2019dvc,neurips2021dcvc,cvpr2024fm,cvpr2025rt}, and the new information of a frame is carried by a latent representation.

Many video applications require high reconstruction fidelity. YCbCr 4:2:0 is primarily the format of distribution to consumers, where the bitrate is the binding constraint and codecs are typically operated at moderate quality. High fidelity is required mainly earlier in the chain, in content acquisition, postproduction, contribution, and archiving, where video is commonly carried as 10-bit YCbCr 4:2:2~\cite{tcsvt2016rext} and production codecs such as ProRes~\cite{apple2022prores} and APV~\cite{rfc9924apv} define the qualities at which it is handled. However, most learned video codecs do not target this range. Recent models widen the quality range of a single model~\cite{cvpr2024fm} and report a loss of efficiency toward high quality~\cite{cvpr2025rt,cvpr2026uf}.

How much of this quality range does a learned codec cover? We bound the range by two established references: the lowest quality at which video coding standardization evaluates codecs, VVC at QP 37, and the highest quality at which production handles 4:2:2 video, the HQ grade of ProRes. To compare inter-frame coding alone, every codec in this paper receives original frames as references at the intra positions. Even so, four recent learned codecs with public models cover 24 to 38\% of the range on the JVET 4:2:2 test sequences (Fig.~\ref{fig:teaser}).

Our measurements point to the structure of these codecs. They encode a frame in one forward pass through networks whose latent capacity, the number of latent values per sample, is fixed and at most 0.25. Removing the quantization of their latent values adds at most 1\,dB, in line with the known limit of dimension-reducing transforms in learned image coding~\cite{iclrw2021flows,acmmm2021eie}. What the latent cannot carry must come from the temporal prediction, and the quality they reach varies with how well a motion-compensated prediction explains the same frames. That prediction is obtained in one pass as well, from a flow that a network estimates, and in multi-reference prediction from distant frames it is weaker than block motion searched by an encoder.

Conventional codecs handle both by encoder search: the encoder tries candidates and keeps the one with the lowest rate-distortion cost for the content at hand. A feed-forward encoder instead applies one mapping that training has made good on average over its data, so individual contents are served worse than they could be, a difference known as the amortization gap~\cite{neurips2020sga}. Search optimizes the decisions for each frame and block at encoding time. To this end, we propose a learned inter-frame codec with encoder search. It searches the motion: its prediction follows the inter-prediction design of VVC, with encoder-searched block motion, compact motion syntax, and a learned refinement that needs no side information. It also searches the transform: the residual is coded by learned transforms on blocks from 4$\times$4 to 64$\times$64, whose latent capacity ranges from 1 for the smallest blocks to 0.016 for the largest, and a candidate search with a rate-distortion cost selects the partition, the skip decisions, and the quantized coefficients. Our codec codes 10-bit YCbCr 4:2:2 natively.

Our codec reaches the upper end of the range on all six sequences. Two ablations show what the search provides. With the block size fixed so that the latent capacity is 0.25, the same codec stops at 39.7\,dB, with or without quantization, and covers 55\% of the range, and with the capacity fixed to 1 it needs more bitrate at every quality. With distant references, the searched motion gives a better predictor than the transmitted flow at a comparable motion rate. The search costs encoding time, and our codec does not yet match the rate efficiency of HEVC and VVC (Sec.~\ref{sec:experiments}). Our contributions are as follows:

\begin{itemize}
 \item We measure the quality range that learned video codecs cover and relate its limit to their fixed latent capacity and to the quality of the temporal prediction.
 \item We propose a learned inter-frame codec for 10-bit YCbCr 4:2:2 whose encoder searches block motion and searches among learned block transforms with latent capacities up to 1.
 \item We show that our codec reaches the upper end of the range on all six JVET 4:2:2 test sequences, and ablations confirm what each search provides.
\end{itemize}

\section{Related Work}
\label{sec:related}

\begin{figure*}[t]
\centering
\includegraphics[width=0.9\textwidth]{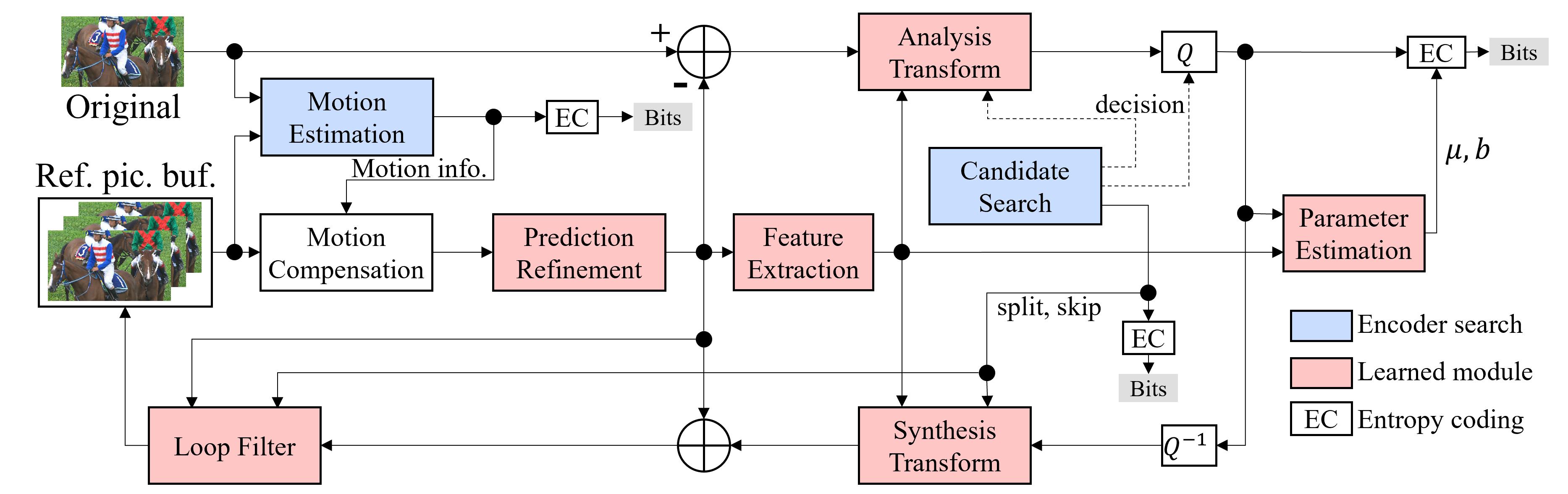}
\caption{Framework of the proposed codec. Blue blocks are searched by the encoder, pink blocks are learned, and the remaining blocks are fixed operations. Dashed lines are decisions of the candidate search. $Q$ and $Q^{-1}$ denote quantization and dequantization, EC entropy coding, and $\mu$ and $b$ the parameters of the Laplace model. The decoder consists of all blocks except motion estimation, the candidate search, the analysis transform, and $Q$.}
\label{fig:framework}
\end{figure*}

\paragraph{Learned video compression.}
DVC~\cite{cvpr2019dvc} replaced the modules of a hybrid codec with networks and coded a motion latent and a residual latent. Conditional coding~\cite{neurips2021dcvc} supplies temporal context to the transforms and the entropy model instead of subtracting a prediction, and its successors widened the quality range of a single model~\cite{cvpr2024fm}, removed explicit motion~\cite{cvpr2025rt}, and reduced complexity~\cite{cvpr2026uf}. Bidirectional models code hierarchical B-frames from past and future references~\cite{tcsvt2023bcanf,tcsvt2024maskcrt,tmm2025dcvcb,arxiv2025biecvc,neurips2025brhvc,arxiv2026bicrvc}.

\paragraph{Encoder-side search.}
Learned codecs usually encode in one forward pass, which is optimal on average over the training data but not for an individual content~\cite{neurips2020sga}. Encoder-side search has entered learned compression to close this gap, as the refinement of the latent of an image~\cite{cvprw2019contentadaptive,neurips2020sga}, as block-level mode selection~\cite{cvpr2023mmvc}, and as block-based motion search that supplies a learned codec~\cite{icip2023blockme}. Our codec searches both the motion and the residual coding: it combines explicit motion syntax with a candidate search over the partition and the coefficients of learned block transforms.

\paragraph{Temporal alignment.}
Most learned codecs obtain their temporal prediction from an optical flow that a network estimates and that is transmitted as a latent~\cite{cvpr2019dvc,neurips2021dcvc}. Others model the temporal relation without explicit motion~\cite{cvpr2025rt} or code B-frames without transmitting motion~\cite{cvpr2023tlzmc}. Bidirectional models align two references over the frame distances of a hierarchical GOP~\cite{tmm2025dcvcb,neurips2025brhvc}. We compare transmitted flow with searched block motion under a common compensation and use the latter.

\paragraph{Quality range and invertible transforms.}
Analysis and synthesis transforms that reduce dimensionality discard information before quantization, which limits the quality range a model covers. Invertible transforms were proposed for learned image coding to remove this limit~\cite{iclrw2021flows,acmmm2021eie}, and were applied to video~\cite{tcsvt2023bcanf,tbc2025invertiblevideo}. Learned wavelet video coding supports operation up to lossless reconstruction~\cite{arxiv2024pmctf}. Invertible transforms have also been combined with implicit conditioning to target high fidelity in video coding~\cite{arxiv2026innvc}. Our codec reaches high quality through small blocks whose transforms are selected by the encoder.

\paragraph{Conventional and professional codecs.}
The range extensions of HEVC~\cite{tcsvt2016rext} and the corresponding VVC profiles~\cite{tcsvt2021vvcoverview} support 4:2:2 and 4:4:4 at high bit depths with inter prediction, and JVET maintains test conditions for these formats~\cite{jvet2020non420}. ProRes~\cite{apple2022prores} and APV~\cite{rfc9924apv} are codecs for acquisition and editing. Learned image coding has addressed subsampled chroma formats, including 4:2:2~\cite{ojsp2021subsampled,smpte2022chroma,smptej2023chroma,tcsvt2025jpegai}. Neural network-based tools have also been integrated into VVC as in-loop filters and intra prediction~\cite{arxiv2023nnvc}, and our in-loop filter follows this line.

\section{Proposed Method}
\label{sec:method}

\subsection{Overview}
\label{sec:overview}

Figure~\ref{fig:framework} shows the proposed codec. It is a hybrid coding loop for hierarchical B-frames in a GOP of 32 with six temporal layers. Its encoder searches at the two blue blocks of the figure: motion estimation searches block motion, and the candidate search selects the residual partition, the skip decisions, and the quantized coefficients. All decisions are signalled explicitly. Networks refine the prediction, transform and model the residual, and filter the reconstruction. We adopt the inter prediction and the entropy coding of syntax elements from VVC~\cite{tcsvt2021vvcoverview} and the structure of the loop filter from NNVC~\cite{arxiv2023nnvc}. We propose the learned block transforms with latent capacities up to 1, their entropy model, and the candidate search over them. Frames are 10-bit YCbCr 4:2:2 throughout, and no stage converts them to another format. Intra frames are supplied from outside the codec, and B-frames contain no intra-coded blocks.

\subsection{Prediction}
\label{sec:prediction}

\paragraph{Motion search and compensation.}
The prediction follows the inter-prediction design of VVC~\cite{tcsvt2021vvcmv,tcsvt2021vvcsubblock} with a subset of its tools, which the supplementary material lists. Motion units form a quadtree from 64$\times$64 down to 4$\times$4. A unit is coded in merge mode, with spatial, temporal, and history-based candidates, or with a motion vector difference at quarter-, integer-, or four-pel resolution, and it is predicted from one or two of up to four reference frames per list. Larger units may use a four-parameter affine model. The encoder selects the partition and the modes by minimizing a prediction error plus $\lambda$ times the motion rate, following the motion search of the VTM encoder, and the motion syntax is coded with the CABAC engine of VVC~\cite{tcsvt2003cabac,tcsvt2021vvcquant}. The prediction is formed with the interpolation filters of VVC and its decoder-side tools DMVR, BDOF, and PROF, which improve the prediction without motion bits.

\paragraph{Learned refinement.}
We refine the motion-compensated frame with a network that outputs an additive correction for all three components. Its luma branch is a U-Net on the half-resolution grid, and its chroma branch operates on the quarter-resolution grid and receives the luma features. The network takes the temporal layer and the operating point as further inputs, has 0.72\,M parameters, and uses no side information. The refined prediction $\hat{x}$ is used by all later stages.

\subsection{Learned block transforms}
\label{sec:residual}

\paragraph{Latent capacity.}
We code the residual $r = x - \hat{x}$ on square blocks of $N\times N$ luma samples, $N \in \{4, 8, 16, 32, 64\}$, which form a quadtree that is independent of the motion partition. A block holds $2N^2$ samples in 4:2:2, and each block size has its own analysis and synthesis transform with $L$ latent values. We define the latent capacity of a transform as $c = L/(2N^2)$. The transforms for 4$\times$4 and 8$\times$8 blocks have $L=32$ and $L=128$, a latent capacity of 1, and those for the larger blocks have $L=128$, capacities of 0.25, 0.063, and 0.016. The capacity is an upper limit, and how many of these values a trained transform uses depends on its training (Sec.~\ref{sec:training}). Both transforms consist of three $1\times1$ convolutions on the grid of blocks, so each block is transformed and reconstructed independently of its neighbors.

\paragraph{Conditioning on the prediction.}
The transforms follow conditional residual coding~\cite{tcsvt2024crc,tcsvt2024maskcrt}: features of the refined prediction, extracted by $3\times3$ convolutions on the block grid, are added inside the analysis and synthesis transforms and supplied to the entropy model. The temporal layer and the quantization step enter through adaptive feature modulation (AFM)~\cite{arxiv2022afm}, a channel-wise scaling and offset $h \leftarrow h\,(1+a)+c$ whose parameters $a$ and $c$ are a linear function of the two inputs, as in other feature-wise conditioning~\cite{aaai2018film,cvpr2024fm}. AFM layers follow the convolutions of the feature extraction and the first two layers of each transform. They tell the transforms the temporal layer of the frame, because the quantization step varies with the temporal layer within one operating point, whereas the gains and the probability tables are defined per operating point. The prediction features are the only path through which a block depends on its neighborhood, so the reconstruction of a block does not depend on the coefficients of other blocks. The 64$\times$64 transform is not conditioned.

\paragraph{Quantization and entropy coding.}
Learned gains per channel and per operating point scale the latent before quantization and after dequantization, as in variable-rate learned codecs~\cite{iccv2019condae,cvpr2021gain,icip2023qvrf,mm2022dcvchem,cvpr2023dcvcdc,cvpr2024fm,cvpr2025rt,cvpr2026uf}. The scaled latent values are quantized uniformly. Our codec has eight operating points, each with its own quantization step, and the step grows with the temporal layer of the frame. We label the operating points of our codec by a pseudo-QP, pQP~$N$: $N$ is the QP of the VTM anchor whose quality range the setting was designed for, not a quantization parameter of our codec. Each quantized value is modeled by a Laplace distribution with mean $\mu$ and scale $b$, as in mean-scale entropy models~\cite{iclr2018hyperprior,neurips2018joint}. Our coder has no hyperprior and no context network. Instead, $\mu$ and $b$ are read from a learned table indexed by a context, which is formed from the magnitudes of the values already coded in the same block and in the left and upper blocks, in the spirit of the context modeling of transform coefficient levels in VVC~\cite{tcsvt2021vvcquant}, and the prediction features correct the table values. The values are coded with a range asymmetric numeral system (rANS) coder~\cite{arxiv2013ans,arxiv2014interleaved}, with probabilities read from tables indexed by the quantized scale~\cite{arxiv2020compressai,mm2022dcvchem,cvpr2023dcvcdc,cvpr2024fm,cvpr2025rt}. A value whose predicted scale lies below a threshold is not coded and is set to its predicted mean at both encoder and decoder~\cite{eccv2022alphavc,cvpr2025rt,cvpr2026uf}. Split and skip flags of the residual quadtree are coded with the CABAC engine of VVC.

\subsection{Candidate search}
\label{sec:search}

For every frame, the encoder selects the quantized values, the skip decisions, and the partition of the residual by minimizing
\begin{equation}
 J = D + \lambda R,
 \label{eq:cost}
\end{equation}
where $D$ is the weighted squared error of the three components, $R$ the rate of coefficients and flags, and $\lambda$ is set by the operating point. First, the quantized latent of each block is refined in the manner of rate-distortion optimized quantization~\cite{vceg2008rdoq}, which the HEVC and VVC reference encoders use and which corresponds to the encoder-side latent refinement of learned image codecs~\cite{cvprw2019contentadaptive,neurips2020sga}: for each non-zero value in coding order, keeping it, moving it one step toward zero, and setting it to zero are compared by their actual cost, with the synthesis transform evaluated for each candidate. Then a bottom-up pass over the quadtree chooses for every node
\begin{equation}
 m^{\ast} = \arg\min_{m \,\in\, \{\mathrm{code},\,\mathrm{skip},\,\mathrm{split}\}} J(m).
 \label{eq:mode}
\end{equation}
Because synthesis is local to a block, the candidates of all blocks of a size are evaluated as one batch on the GPU, which makes this search practical.

\subsection{In-loop filtering}
\label{sec:loopfilter}

We derive the loop filter from the low-complexity in-loop filter of NNVC~\cite{arxiv2023nnvc}, replace its dynamic convolutions by ordinary ones, and adapt the inputs and outputs to 4:2:2. It receives the reconstruction, the prediction, block-boundary maps derived from the residual partition and its skip decisions, and the quantization step. It has 0.24\,M parameters and needs no side information.

\subsection{Training}
\label{sec:training}

We train on clips of 33 frames from two sources. The first are 2{,}590 clips cut from full-length sources that are available as ProRes 4:2:2. The second are 160 sequences of the BVI-AOM dataset~\cite{arxiv2024bviaom}, converted from 1920$\times$1088 4:2:0 to 960$\times$544 4:2:2. No test sequence is part of the training data. Motion search is not differentiable, so the motion decisions are computed beforehand and kept fixed. Training has three steps. The block transforms are first trained separately for each block size with the rate-distortion cost of Eq.~(\ref{eq:cost}) as the loss. Then the refinement network, the five block transforms with their entropy models, and the loop filter are trained jointly on chains of seven frames, two original frames 32 frames apart and one frame per temporal layer between them, so that gradients pass through the reference frames. Finally, the block transforms are trained once more for all eight operating points with the other networks kept. The distortion weights luma, Cb, and Cr as 2:1:1, the sample counts of 4:2:2, and the candidate search uses the same weights.

The highest operating point, pQP~0, has half the quantization step of pQP~7, and at this point all latent values of the 4$\times$4 and 8$\times$8 transforms are in use. For pQP~7 and pQP~0, Gaussian noise augmentation is applied to the chroma of the training samples. The supplementary material gives further training details.

\section{Experiments}
\label{sec:experiments}

\subsection{Experimental settings}
\label{sec:setup}

\paragraph{Test sequences and metrics.}
We use the 10-bit YCbCr 4:2:2 sequences of the JVET test set for non-4:2:0 formats~\cite{jvet2020non420}: EBUHorse, EBUKidsSoccer, EBUWaterRocksClose, and Seeking (1920$\times$1080, 50\,fps), Kimono1 (1920$\times$1080, 24\,fps), and Traffic (2560$\times$1600, 30\,fps). The seventh sequence of the set, EBUGraphics, is screen content and is outside our scope. Rates are the sizes of the actual bitstreams in bits per pixel (bpp), counting the luma samples of the coded frames. Quality is the PSNR of the squared error averaged over all samples of a frame, which in 4:2:2 weights Y, Cb, and Cr as 2:1:1. This is the combined PSNR of the VVC reference software.

\paragraph{Test condition.}
This paper compares inter-frame coding. The codecs we compare do not share an intra coder, and the quality of the intra frames changes what the following frames can reach. To isolate inter coding, original frames are placed at the intra positions at zero rate, and rate and PSNR are those of the coded frames only. All codec comparisons of the main paper are made in this condition. The bidirectional codecs, including ours, receive the originals of frames 0 and 32 and code frames 1 to 31 as hierarchical B-frames. The unidirectional codecs receive frame 0.

\paragraph{Compared codecs.}
We compare with the bidirectional BRHVC~\cite{neurips2025brhvc} and DCVC-B~\cite{tmm2025dcvcb}, the unidirectional DCVC-RT~\cite{cvpr2025rt}, and the chunk-based model of DCVC-UF~\cite{cvpr2026uf}, all run with their public code and weights at quality settings spanning their official range. They operate internally on 4:4:4, so the 4:2:2 chroma is replicated horizontally before coding and averaged back afterwards. For HEVC and VVC we run HM 19.0~\cite{softwarehm19} and VTM 23.14~\cite{software2026vtm2314} in their random-access configurations for 10-bit 4:2:2 with an intra period of 32 at QP 12 to 37, and HM also at QP 7 on Kimono1 and Traffic. Our codec is run at eight operating points, pQP~37 to pQP~7 in steps of 5 and pQP~0 (Sec.~\ref{sec:residual}), and every point is decoded by a separate decoder whose output matches the encoder reconstruction. BD-rates are computed with piecewise cubic interpolation of the logarithmic rate over the overlapping quality interval and averaged over sequences.

\begin{figure*}[t]
 \centering
 \includegraphics[width=\textwidth]{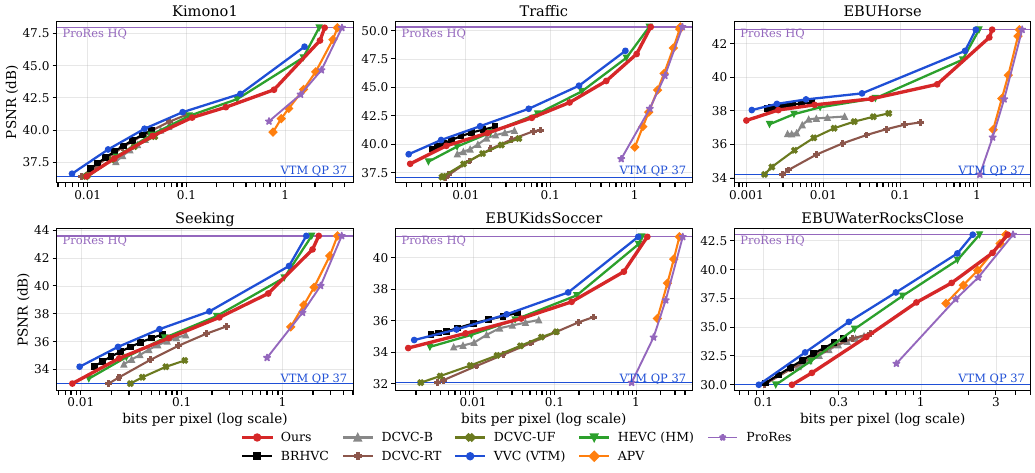}
 \caption{Rate-distortion curves of the inter frames on each sequence, within the comparison range from the quality of VVC at QP 37 to that of ProRes HQ (horizontal lines). The rate axis is logarithmic and counts the inter frames only. Our codec, the four learned codecs, HEVC, and VVC use original frames at the intra positions (Sec.~\ref{sec:setup}). APV and ProRes are shown for reference.}
 \label{fig:range}
\end{figure*}

\subsection{Comparison with learned codecs}
\label{sec:range}

\begin{table}[t]
 \centering
 \footnotesize
 \setlength{\tabcolsep}{3.2pt}
 \begin{tabular}{@{}lrrrrrrr@{}}
 \toprule
 & & \multicolumn{3}{c}{DCVC} & \multicolumn{3}{c}{Ours, latent capacity} \\
 \cmidrule(lr){3-5}\cmidrule(l){6-8}
 Sequence & BRHVC & B & RT & UF & adaptive & 1 & 0.25 \\
 \midrule
 Kimono1 & 31 & 34 & 38 & 27 & 100 & 100 & 61 \\
 Traffic & 34 & 31 & 31 & 26 & 100 & 100 & 62 \\
 EBUHorse & 50 & 40 & 35 & 42 & 100 & 100 & 69 \\
 Seeking & 33 & 33 & 35 & 16 & 100 & 100 & 57 \\
 EBUKidsSoccer & 48 & 43 & 42 & 35 & 100 & 100 & 60 \\
 EBUWaterRocksClose & 31 & 32 & 32 & 0 & 100 & 92 & 23 \\
 \midrule
 Average & 38 & 35 & 36 & 24 & 100 & 99 & 55 \\
 Without quantization & 43 & 43 & 45 & 28 & 100 & 98 & 55 \\
 \bottomrule
 \end{tabular}
 \caption{Share (\%) of the quality range from VTM at QP 37 to the high-quality grade of ProRes that each codec reaches at its highest setting. The last row is the average share when the rounding of the latent values is removed, which is not an upper bound. All codecs receive original frames at the intra positions. Latent capacity 1 and 0.25 are our codec with all residual blocks 8$\times$8 and 16$\times$16.}
 \label{tab:coverage}
\end{table}

\paragraph{Comparison range.}
For about two decades, from the common conditions of H.264/AVC experiments~\cite{vceg2005common} to the common test conditions of VVC~\cite{jvet2020ctc}, video coding standardization has compared coding tools at four quantization parameters, QP 22, 27, 32, and 37. We take the lowest of these qualities as the lower end of the comparison range: per sequence, the quality of the inter frames of the VTM random-access anchor at QP 37, where VTM codes its own intra frames. The highest of these qualities, QP 22, covers only 53\% of the range that we use, so we take the upper end from production: the quality of the HQ grade of ProRes~\cite{apple2022prores} produced by the Apple encoder, which lies between 41.3 and 50.3\,dB. The share that a codec covers is the highest quality it reaches minus the lower end, divided by the width of the range and limited to 100\%.

\paragraph{Results.}
Table~\ref{tab:coverage} gives the shares, and Fig.~\ref{fig:range} shows the rate-distortion curves within the range. The four learned codecs cover 24 to 38\% of the range on average, and their share varies with the content, from 0 to 50\% across sequences. Their highest quality settings end at 37.8\,dB for BRHVC, 37.6\,dB for DCVC-B, 37.7\,dB for DCVC-RT, and 36.0\,dB for DCVC-UF. Our codec covers 34.3 to 58.9\,dB with rates from 0.015 to 11.2 bpp and reaches the upper end of the range on all six sequences. It reaches the quality of ProRes HQ at 1.4 to 3.6 bpp, found by interpolation between the two neighboring operating points. This quality lies at or below pQP~7 on four sequences and between pQP~7 and pQP~0 on Kimono1 and Traffic. Within the range that the learned codecs cover, BRHVC is the most efficient: between 35.7 and 37.8\,dB it needs 34 to 36\% fewer bits than our codec. Our codec needs fewer bits than the other three over the range it shares with them.

\subsection{Comparison with HEVC and VVC}
\label{sec:standard}

\begin{table}[t]
 \centering
 \small
 \setlength{\tabcolsep}{4pt}
 \begin{tabular}{@{}lrr@{}}
 \toprule
 Sequence & HEVC & VVC \\
 \midrule
 Kimono1 & $+15.7$ & $+59.1$ \\
 Traffic & $+6.8$ & $+81.1$ \\
 EBUHorse & $+7.4$ & $+129.4$ \\
 Seeking & $+4.9$ & $+57.4$ \\
 EBUKidsSoccer & $+14.1$ & $+87.1$ \\
 EBUWaterRocksClose & $+41.0$ & $+71.9$ \\
 \midrule
 Average & $+15.0$ & $+81.0$ \\
 \bottomrule
 \end{tabular}
 \caption{BD-rate (\%) of our codec against HEVC (HM 19.0) and VVC (VTM 23.14) over the comparison range.}
 \label{tab:bdseq}
\end{table}

Table~\ref{tab:bdseq} gives the BD-rates of our codec against HEVC and VVC over the comparison range, $+15.0\%$ and $+81.0\%$ on average. HEVC itself requires 52.6\% more bitrate than VVC. The prediction of our codec omits several inter tools of VVC, among them binary and ternary splits, MMVD, geometric partitioning, and BCW, and has no intra-coded blocks in B-frames.

\subsection{Ablation study}
\label{sec:fixedsize}

\begin{table}[t]
 \centering
 \footnotesize
 \setlength{\tabcolsep}{2.6pt}
 \begin{tabular}{@{}lrrrrrr@{}}
 \toprule
 Blocks & Capacity & Highest & No quant. & Share & Reached & BD-rate \\
 \midrule
 Adaptive & 0.016 to 1 & 58.9 & 59.3 & 100 & 6/6 & 0 \\
 All 8$\times$8 & 1 & 48.8 & 48.7 & 99 & 5/6 & $+18.9$ \\
 All 16$\times$16 & 0.25 & 39.7 & 39.7 & 55 & 0/6 & $+60.5$ \\
 \bottomrule
 \end{tabular}
 \caption{Our codec with the residual block size fixed: latent capacity, highest PSNR (dB), PSNR with the rounding of the latent values removed (dB), covered share of the range (\%), number of sequences on which the upper end is reached, and BD-rate (\%) against the adaptive codec over the part of the range that the configuration covers.}
 \label{tab:ablation}
\end{table}

\begin{figure}[t]
 \centering
 \includegraphics[width=\linewidth]{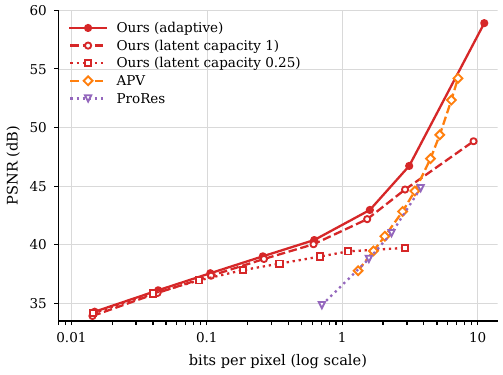}
 \caption{Our codec with the residual block size selected by the encoder (adaptive), fixed to 8$\times$8 (latent capacity 1), and fixed to 16$\times$16 (latent capacity 0.25). APV at QP 20 to 40 and the four 4:2:2 grades of ProRes are shown for reference.}
 \label{fig:learned}
\end{figure}

\paragraph{Latent capacity.}
To test the role of the small blocks and of their latent capacity, we run our codec with the residual block size fixed and nothing else changed. The weights are those trained with the adaptive partition, and the split flags are still sent and counted. Table~\ref{tab:ablation} and Fig.~\ref{fig:learned} give the result. With all blocks 16$\times$16, whose latent capacity is 0.25 as in the other learned codecs, the codec ends at 39.7\,dB and covers 55\% of the range. Removing the quantization does not change this: with the rounding of its latent values removed, the same configuration stays at 39.7\,dB. With all blocks 8$\times$8, a latent capacity of 1, it reaches the upper end on five sequences but needs 18.9\% more bitrate than the adaptive codec, about 20\% in the lower and in the upper half of the range alike. The configuration with all blocks 16$\times$16 needs 60.5\% more bitrate over the part of the range that it covers. The adaptive codec is thus more efficient than either fixed capacity at every quality. The range is therefore obtained by the small blocks, whose transforms do not reduce the dimension, and the selection of the block size lowers the rate. The capacity is an upper limit and not a guarantee: the 8$\times$8 transform has a capacity of 1 like the 4$\times$4 transform but ends at 48.8\,dB, whereas the adaptive codec reaches 58.9\,dB with 4$\times$4 blocks. Since the configurations differ in block size and transform as well as in capacity, the test does not isolate the capacity alone. The adaptive codec spans the range by changing its partition. At pQP~37, 64$\times$64 blocks cover 84\% of the frame area and the residual of 95\% of the area is skipped. The share of the small blocks grows with the quality, and at pQP~0 the whole frame is coded with 4$\times$4 blocks, so the latent capacity of the blocks in use moves from 0.016 to 1. The partition also follows the content: at the same operating point, pQP~22, 64$\times$64 blocks cover from 18\% of the frame area on Kimono1 to 76\% on EBUHorse, and the residual is skipped on 47 to 96\% of the area. The search thus sets the latent capacity per block and per content, which a fixed transform cannot do.

\begin{table}[t]
 \centering
 \small
 \setlength{\tabcolsep}{4pt}
 \begin{tabular}{@{}lrrr@{}}
 \toprule
 Codec & Highest & No quant. & Capacity \\
 \midrule
 BRHVC & 37.8 & 38.4 & 0.25 \\
 DCVC-B & 37.6 & 38.4 & 0.25 \\
 DCVC-RT & 37.7 & 38.7 & 0.25 \\
 DCVC-UF & 36.0 & 36.4 & 0.06 \\
 \bottomrule
 \end{tabular}
 \caption{PSNR (dB) of the four learned codecs at their highest quality setting (``Highest'') and with the rounding of their latent values removed (``No quant.''), and their latent capacity. Six-sequence averages.}
 \label{tab:ceiling}
\end{table}

\paragraph{Quantization.}
The other learned codecs show the same limit. A codec that stops at some quality might simply lack finer quantization settings, so we reconstruct each model at its highest official quality setting with the rounding of its latent values removed. Table~\ref{tab:ceiling} gives the result. The four models gain at most 1.0\,dB and would cover 28 to 45\% of the range (Table~\ref{tab:coverage}). The quality of the released models is therefore not limited by the quantization step. We did not retrain them for higher quality, so this holds for the models as released. BRHVC, DCVC-B, and DCVC-RT transmit, for every frame and at every quality setting, 128 channels at one sixteenth of the resolution in each dimension for the frame content, which is 0.5 values per pixel and, with 2 samples per pixel in 4:2:2, a latent capacity of 0.25. DCVC-UF transmits 256 channels for a chunk of eight frames, a capacity of 0.06. A representation with a capacity below 1 cannot reproduce an arbitrary frame even without quantization~\cite{iclrw2021flows}.

\begin{figure*}[t]
 \centering
 \includegraphics[width=0.7\textwidth]{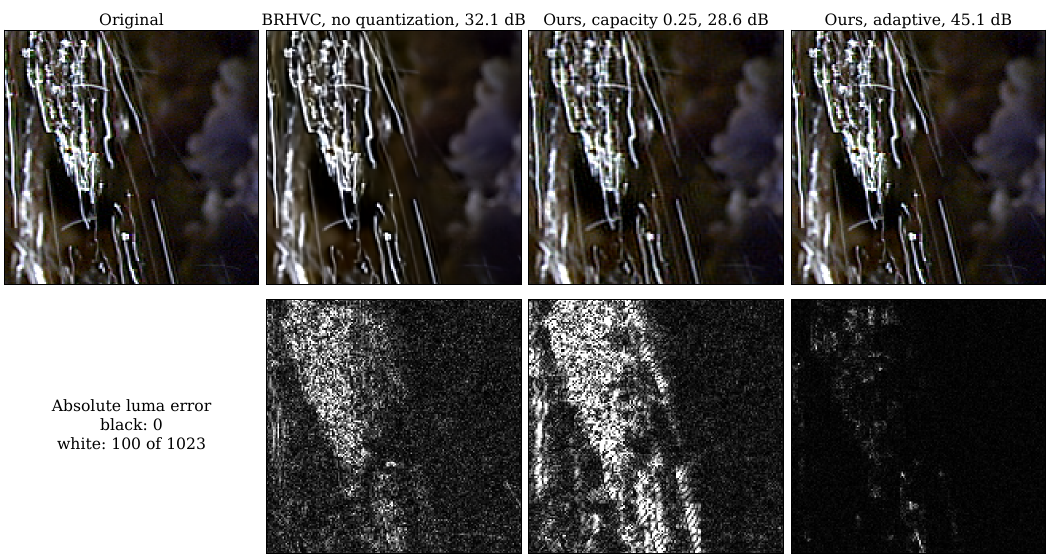}
 \caption{EBUWaterRocksClose, frame 16, a crop of 160$\times$160 luma samples: reconstructions with the PSNR of the crop (top) and their absolute luma error (bottom). BRHVC is reconstructed without quantization, and our codec is at pQP~7.}
 \label{fig:visual}
\end{figure*}

\paragraph{Visual comparison.}
Figure~\ref{fig:visual} shows a crop of EBUWaterRocksClose, the sequence that temporal prediction explains least. BRHVC, even without quantization, restores the coarse shape of the water streaks but not their fine structure, and our codec with a latent capacity of 0.25 behaves in the same way. The error maps show that both lose the same regions. With the adaptive partition, the fine structure is reproduced and the error almost vanishes.

\begin{table}[t]
 \centering
 \footnotesize
 \setlength{\tabcolsep}{2.6pt}
 \begin{tabular}{@{}lrrrrrr@{}}
 \toprule
 & Pred. & \multicolumn{4}{c}{Without quantization} & Ours \\
 Sequence & (ours) & BRHVC & DCVC-B & RT & UF & pQP~0 \\
 \midrule
 Kimono1 & 39.8 & 40.8 & 41.2 & 41.9 & 40.2 & 60.8 \\
 Traffic & 42.3 & 42.2 & 42.0 & 42.5 & 41.2 & 61.3 \\
 EBUHorse & 38.5 & 38.6 & 38.3 & 37.9 & 38.2 & 61.7 \\
 Seeking & 36.3 & 37.0 & 37.2 & 37.8 & 35.0 & 60.5 \\
 EBUKidsSoccer & 36.7 & 36.9 & 36.9 & 37.0 & 35.7 & 61.5 \\
 EBUWaterRocksClose & 25.4 & 34.9 & 34.9 & 34.8 & 28.3 & 47.5 \\
 \midrule
 Average & 36.5 & 38.4 & 38.4 & 38.7 & 36.4 & 58.9 \\
 \bottomrule
 \end{tabular}
 \caption{PSNR (dB) of the temporal predictor of our codec at its highest quality setting, of the four learned codecs reconstructed without quantization, and of our codec at its highest quality setting. RT and UF denote DCVC-RT and DCVC-UF.}
 \label{tab:predceiling}
\end{table}

\begin{table}[t]
 \centering
 \footnotesize
 \setlength{\tabcolsep}{2.1pt}
 \begin{tabular}{@{}lrrrrrrrr@{}}
 \toprule
 & \multicolumn{2}{c}{$d=16$} & \multicolumn{2}{c}{$d=8$} & \multicolumn{2}{c}{$d=4$} & \multicolumn{2}{c}{$d=1$} \\
 Motion source & rate & dB & rate & dB & rate & dB & rate & dB \\
 \midrule
 Zero motion & 0 & 22.9 & 0 & 24.0 & 0 & 25.0 & 0 & 29.5 \\
 Flow of BRHVC, lowest & 9.8 & 26.3 & 8.8 & 28.6 & 7.1 & 31.4 & 3.4 & 34.7 \\
 Flow of BRHVC, highest & 15.6 & 26.4 & 14.7 & 28.7 & 12.5 & 31.6 & 6.6 & 34.9 \\
 Searched, pQP~37 & 16.5 & 31.7 & 14.3 & 32.6 & 5.5 & 31.5 & 1.9 & 32.9 \\
 Searched, pQP~7 & 214 & 34.9 & 169 & 35.5 & 82.8 & 35.0 & 20.2 & 34.5 \\
 \bottomrule
 \end{tabular}
 \caption{Quality of the predictor that each motion source gives under a common compensation operator, with original reference frames at distance $d$ on both sides, averaged over the six test sequences. Rate is the motion rate in $10^{-3}$ bpp. Quality is the PSNR over all samples of the three components.}
 \label{tab:align422}
\end{table}

\paragraph{Temporal prediction.}
If the latent cannot carry the whole frame, the rest must come from the temporal prediction. Table~\ref{tab:predceiling} compares the quality that each learned codec reaches without quantization with the PSNR of the temporal predictor of our codec at its highest setting, used as a proxy for a motion-compensated prediction of the same frames. On the five sequences other than EBUWaterRocksClose, the unquantized quality of the four models lies between 1.3\,dB below and 2.1\,dB above the predictor. On EBUWaterRocksClose, a water surface that temporal prediction hardly explains, the predictor reaches 25.4\,dB and the four models end between 28.3 and 34.9\,dB, the lowest values of all sequences. On these five sequences the quality of the released models thus stays close to that of the predictor, whereas on EBUWaterRocksClose three of them end 9.4 to 9.5\,dB above it. Our codec reaches 19.0 to 24.8\,dB above its own predictor, whereas with all blocks 16$\times$16 it ends 0.9 to 3.5\,dB above it on the five sequences. With a latent capacity of 1, the prediction no longer sets the reachable quality. It determines the rate at which a quality is reached.

\paragraph{Motion source.}
\label{sec:alignment422}
We therefore compare the flow that a learned codec transmits, taking BRHVC, with the block motion that our encoder searches. For a target frame and two original reference frames at distance $d$ on either side, the decoded motion of each method is applied to the same references with the same operator: bilinear sampling and, where both references are used, the equal-weight average of the two compensated frames, with the horizontal component halved for chroma. No learned stage and no codec-specific interpolation is involved, so the motion is the only difference. Table~\ref{tab:align422} gives the averages over the six sequences, with one target frame per sequence and distance. With references 16 frames away, the searched motion of pQP~37 gives a predictor 5.3\,dB better than the flow of BRHVC at its highest setting, at nearly the same average motion rate. The difference is 3.9\,dB at $d=8$. At $d=4$ the two are equal, and at adjacent frames the flow gives the better predictor, by 2.0\,dB. The advantage of the search thus lies at long distances, and it varies with the content, from 0.4 to 8.4\,dB at $d=16$. The predictor of the flow changes by at most 0.2\,dB between the lowest and the highest setting of BRHVC, whereas that of the searched motion improves by 1.6 to 3.5\,dB as its rate increases. With the interpolation filters of VVC, its decoder-side refinement, and the learned refinement, the predictor of our codec at $d=16$ reaches 32.7 and 35.6\,dB, 1.0 and 0.7\,dB above the values of the table. The comparison concerns predictors only, because BRHVC compensates features and corrects them with further networks.

\subsection{Complexity}
\label{sec:limits}

On one RTX 6000 Ada GPU, encoding a 1080p frame takes 2 to 8.5\,s and decoding takes 0.4 to 1.2\,s, with less than 5\,GiB of GPU memory. The refinement network, the loop filter, and the five block coders have 0.72, 0.24, and 14.9\,M parameters.

\section{Conclusion and Limitations}
\label{sec:conclusion}

We present a learned inter-frame codec with encoder search for 10-bit YCbCr 4:2:2 that reaches the upper end of a professional quality range, from VVC at QP 37 to the highest 4:2:2 grade of ProRes, on all six test sequences, where four recent learned codecs cover 24 to 38\% of the range. Their latent capacity is at most 0.25, and removing their quantization does not lift the limit. With all residual blocks fixed to 16$\times$16, whose latent capacity is 0.25, our codec covers 55\% of the range. The search over block sizes thus provides the range through the small blocks, whose transforms do not reduce the dimension, and lowers the rate, and the search over motion provides a stronger prediction from distant references.

\paragraph{Limitations.}
The codec is less efficient than HEVC and VVC, and less efficient than BRHVC below 38\,dB. Encoding takes seconds per frame, intra frames are supplied from outside the codec, and the evaluation covers the first 33 frames of six sequences. The other learned codecs are evaluated with their released models, which were not retrained for high quality, and the ablation changes block size and latent capacity together. Closing the rate gap is our next step.

{
    \small
    \bibliographystyle{ieeenat_fullname}
    \bibliography{main}
}

\end{document}